\documentclass[a4paper, 10pt, conference]{IEEEtran}

\pdfoutput=1    

\usepackage{xcolor}
\definecolor{wong-black}        {HTML}{000000}
\definecolor{wong-lightorange}  {HTML}{E69F00}
\definecolor{wong-lightblue}    {HTML}{56B4E9}
\definecolor{wong-green}        {HTML}{009E73}
\definecolor{wong-yellow}       {HTML}{F0E442}
\definecolor{wong-darkblue}     {HTML}{0072B2}
\definecolor{wong-darkorange}   {HTML}{D55E00}
\definecolor{wong-pink}         {HTML}{CC79A7}

\usepackage{url}

\usepackage{hyperref} 
\hypersetup{
    colorlinks=true,
    citecolor=wong-green,
    linkcolor=wong-darkblue,
    filecolor=wong-pink,      
    urlcolor=wong-black,
    pdfpagemode=FullScreen,
    }

\usepackage{cite}
\usepackage{amsmath,amssymb,amsfonts}
\usepackage{algorithmic}
\usepackage{graphicx}
\usepackage{textcomp}
\usepackage[nolist, nohyperlinks, printonlyused]{acronym} 
\usepackage{array,booktabs}
\newcolumntype{M}[1]{>{\centering\arraybackslash}m{#1}} 
\usepackage{subcaption} 

\def\BibTeX{{\rm B\kern-.05em{\sc i\kern-.025em b}\kern-.08em
    T\kern-.1667em\lower.7ex\hbox{E}\kern-.125emX}}
\begin{document}


\title{Multimodal Detection of Unknown Objects\\on Roads for Autonomous Driving}


\author{\IEEEauthorblockN{Daniel Bogdoll\IEEEauthorrefmark{2}\IEEEauthorrefmark{3}\textsuperscript{\textasteriskcentered},
Enrico Eisen\IEEEauthorrefmark{3}\textsuperscript{\textasteriskcentered},
Maximilian Nitsche\IEEEauthorrefmark{3}\textsuperscript{\textasteriskcentered},
Christin Scheib\IEEEauthorrefmark{3}\textsuperscript{\textasteriskcentered},
and J. Marius Zöllner\IEEEauthorrefmark{2}\IEEEauthorrefmark{3}}

\IEEEauthorblockA{\IEEEauthorrefmark{2}FZI Research Center for Information Technology, Germany\\
bogdoll@fzi.de}
\IEEEauthorblockA{\IEEEauthorrefmark{3}Karlsruhe Institute of Technology, Germany\\}}

\maketitle
\begingroup\renewcommand\thefootnote{\textasteriskcentered}
\footnotetext{These authors contributed equally}
\endgroup


\begin{acronym}
    \acro{ml}[ML]{Machine Learning}
    \acro{ODD}[ODD]{Operational Design Domain}
	\acro{cnn}[CNN]{Convolutional Neural Network}
	\acro{dl}[DL]{Deep Learning}
	\acro{ad}[AD]{autonomous driving}
	\acro{DBSCAN}[DBSCAN]{Density-based Spatial Clustering of Applications with Noise}
	\acro{RANSAC}[RANSAC]{Random Sample Consensus}
	\acro{gan}[GAN]{generative adversarial networks}
	\acro{nf}[NF]{normalizing flow}
	\acro{OSIS}[OSIS]{Open-Set Instance Segmentation}
	\acro{av}[AV]{autonomous vehicle}
	\acro{MLUC}[MLUC]{Metric learning with Unsupervised Clustering}
\end{acronym}


\begin{abstract}
Tremendous progress in deep learning over the last years has led towards a future with autonomous vehicles on our roads. Nevertheless, the performance of their perception systems is strongly dependent on the quality of the utilized training data. As these usually only cover a fraction of all object classes an autonomous driving system will face, such systems struggle with handling the unexpected. In order to safely operate on public roads, the identification of objects from unknown classes remains a crucial task. In this paper, we propose a novel pipeline to detect unknown objects. Instead of focusing on a single sensor modality, we make use of lidar and camera data by combining state-of-the art detection models in a sequential manner. We evaluate our approach on the Waymo Open Perception Dataset and point out current research gaps in anomaly detection.
\end{abstract}




\section{Introduction}
\label{sec:introduction}

In December 2021, Mercedes-Benz became the first automotive company to meet the legal requirements for a SAE level 3 system \cite{Daimler_Level3, sae_2018}. However, a driver is still present and must be ready to take over control. This is not the case for level 4 and 5 systems, where the presence of a driver is not required anymore.
Due to significant progress over the last years, multiple level 4 systems are already tested on public roads \cite{Torc_L4,Tusimple_L4,Waymo_L4}. In order for an \ac{av} to appropriately act and react, reliable object detection is crucial. State-of-the-art object detectors are based on \ac{dl} \cite{dibiasePixelwiseAnomalyDetection2021a}. As these models rely on large training datasets, they typically assume that all classes that are to be detected have been present during training \cite{Joseph2021TowardsOW}. If the model encounters instances that are outside the training distribution of the network, also called anomalies or corner cases \cite{Bogdoll_Description_2021_ICCV}, they tend to fail \cite{dibiasePixelwiseAnomalyDetection2021a}. Even by defining an \ac{ODD}, the full set of objects that can occur on roads is not predictable. Thus, a model cannot be trained with all the classes it could potentially face. Nevertheless, in order for an \ac{av} to safely operate, the perception system should detect both, known and unknown classes. This is also referred to as the open-set setting \cite{wongIdentifyingUnknownInstances2019}. 
\begin{figure}[t!]
    \centering
    \begin{subfigure}[t]{0.5\linewidth}
        \centering
        \includegraphics[height=1.1in]{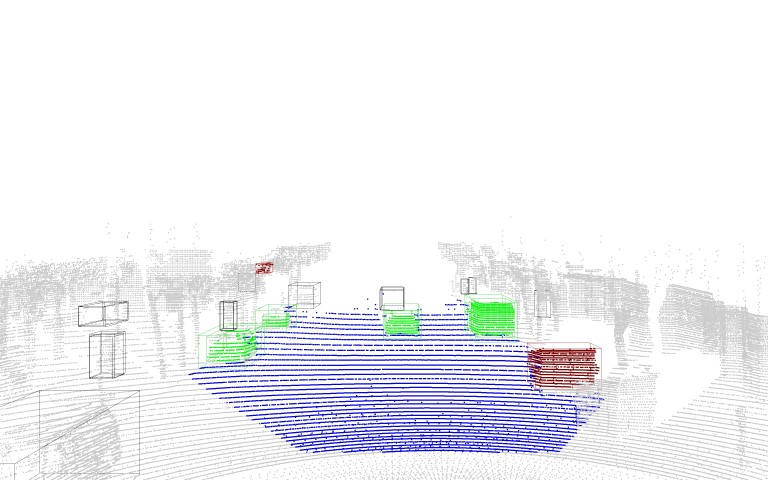}
        \caption{Lidar point cloud}
    \end{subfigure}%
    ~ 
    \begin{subfigure}[t]{0.5\linewidth}
        \centering
        \includegraphics[height=1.1in]{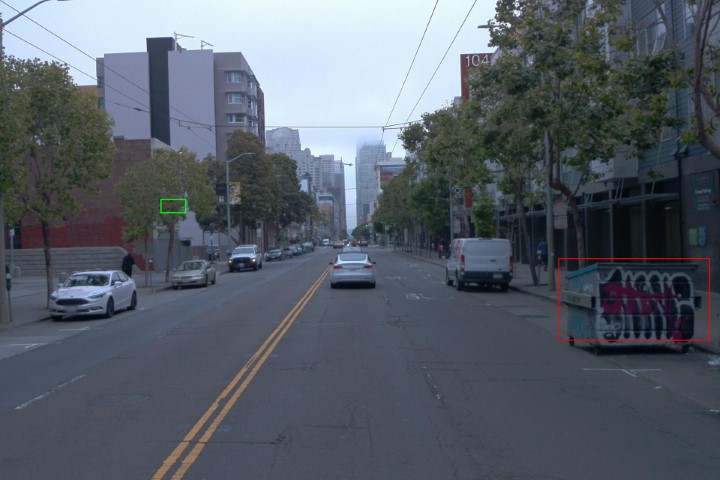}
        \caption{Camera image}
    \end{subfigure}
    \caption{The scene \cite{sunScalabilityPerceptionAutonomous2020} shows, how multimodal anomaly detection is able to detect unknown objects as anomalies. Here, a container is placed on the road. It is detected and classified as unknown in the lidar data (red). Based on the rich semantic information in the camera image, the classifier can infer that it is not a known class and thus an anomaly.}
\end{figure}
A variety of different approaches for that problem exist \cite{dibiasePixelwiseAnomalyDetection2021a,wongIdentifyingUnknownInstances2019,lisDetectingUnexpectedImage2019,nitschOutofDistributionDetectionAutomotive2021}. While AVs consist of a set of different sensors, such as lidars, radars, and cameras, most of the existing work on anomaly detection focuses on one sensor modality \cite{bogdollAnomalyDetectionAutonomous2022}. Hence, only few approaches make use of the advantages of combining sensor modalities. While sensor fusion models are popular for classic object detection, they often lack the awareness necessary for anomaly detection. 
In this work, we suggest a novel pipeline for detecting unknown objects by exploiting the advantages of lidar and camera data. As a definition for anomalies, we follow the definition of object level corner cases by Breitenstein et al.~\cite{9304789}, which describe “instances that have not been seen before”. We limit the problem by focusing on objects on roads, as unusual objects on the sidewalk or in a large distance are less critical for the driving task. We apply semantic segmentation on the input image to derive the road area. We further apply clustering and a 3D object detector on the lidar data. Objects that could not be classified are transformed into the 2D image space and an image classifier is applied. Only when the image classifier is not able to classify the object, we define it as an anomaly. Contrary to sensor fusion approaches, we do not perform simultaneous detection on lidar and camera data. As there is no public anomaly dataset available that consists of more than one sensor modality \cite{Bogdoll_Addatasets_2022_VEHITS}, we use the Waymo Open Perception Dataset \cite{sunScalabilityPerceptionAutonomous2020} for evaluation. To this point, only qualitative evaluation is possible, since we do not have ground truth data available for classes other than vehicles, pedestrians, cyclists, and signs \cite{sun2020scalability}. Our pipeline can be used for offline evaluation of datasets or active learning and has the potential for online applications, such as the activation of remote assistance \cite{Bogdoll_Taxonomy_2022_FICC}. All code is available at \url{https://url.fzi.de/unknown_objects}.



\section{Related Work}
\label{sec:related_work}

Nowadays, existing efforts to detect anomalies in \ac{ad} settings focus mainly on single modality techniques. Especially camera based techniques experience great progress~\cite{bogdollAnomalyDetectionAutonomous2022}, as images provide environmental information with more semantic information compared to lidar and radar.  
The most straightforward approaches to detect anomalies in \ac{ad} are confidence-based. They either utilize the given confidence of the neural network, like the class-confidence, or derive the uncertainty measurement by additional efforts~\cite{kendallBayesianSegNetModel2016, jungStandardizedMaxLogits2021a, heideckerCornerCaseDetection2021, huangEfficientUncertaintyEstimation2018, duVOSLearningWhat2022, chanEntropyMaximizationMeta2021a}. For instance, Kendall~et~al.~\cite{kendallBayesianSegNetModel2016} measure the uncertainty of a semantic segmentation by Monte Carlo dropout sampling. Other efforts focus on learning frameworks that calibrate the network's confidence. These comprise training objectives~\cite{chanEntropyMaximizationMeta2021a}, but especially shaping the decision boundary in a contrastive manner~\cite{grcicDenseAnomalyDetection2021, duVOSLearningWhat2022}. The latter synthesize outliers via sampling a learned distribution in feature space~\cite{duVOSLearningWhat2022} or learning to generate outliers near the decision boundary end-to-end~\cite{grcicDenseAnomalyDetection2021}.
At the end, all the aforementioned approaches determine anomalous objects by thresholding the model's uncertainty. 

On the other hand, reconstructive approaches are well-suited for anomaly detection since they try to reproduce the normality of the training data by design. Anomalous objects are then identified by the difference between the real and re-synthesized image. For instance, Ohgushi~et~al.~\cite{ohgushiRoadObstacleDetection2021} and Voijr~et~al.~\cite{vojirRoadAnomalyDetection2021a} use an autoencoder to re-synthesize the input image where the encoder is part of a semantic segmentation network. The anomaly map is then based on the perceptual loss between the encoder and decoder as well as the entropy loss of the semantic segmentation~\cite{ohgushiRoadObstacleDetection2021}. Similarly, Di~Biase~et~al.~\cite{dibiasePixelwiseAnomalyDetection2021a} refine the image re-synthesis of Lis~et~al.~\cite{lisDetectingUnexpectedImage2019} as they construct the anomaly score by combining entropy, perceptual loss, and the softmax distance. But in contrast to Ohgushi~et~al.'s procedure, their re-synthesis network is based on the final segmentation output. Other approaches utilize generative concepts, like \ac{gan}\cite{nitschOutofDistributionDetectionAutomotive2021, leeSimpleUnifiedFramework2018} and \ac{nf}\cite{grcicDenseAnomalyDetection2021, blumFishyscapesBenchmarkMeasuring2021}, to synthesize anomalous patches or entire driving scenes near the decision boundary between in- and out-of-distribution samples during training. This procedure is motivated to calibrate the model's confidence. 

Besides the surrounding camera system, also other sensor modalities often equipped in \ac{av} can be applicable for anomaly detection. For instance, lidar sensors are often used in \ac{ad} as they provide depth information about the environment. The idea of an open-set detection, however, is only recently adopted to 3D lidar data~\cite{wongIdentifyingUnknownInstances2019, cenOpenset3DObject2021}. Here, Wong~et~al.'s~\ac{OSIS} framework constitutes a baseline for the detection of anomalous objects. The \ac{OSIS} model consists of two branches: a detection and an embedding head. While the former detects the anchors of known classes, the latter learns instance-aware point embeddings and prototypes of the unknown classes. During inference, the prototypes first rule out points under the closed-set condition. The unassigned points are in a second stage clustered via \ac{DBSCAN} into instances of unknown classes and represent the open-set condition.  But in most recent work of Cen~et~al.~\cite{cenOpenset3DObject2021}, the authors show that their \ac{MLUC} significantly outperforms the \ac{OSIS} framework. The metric learning network first obtains classified bounding boxes and their corresponding embeddings. Instances are then again labeled as known in the embedding space whenever they fall in the neighborhood of an embedded prototype. However, in contrast to Wong~et~al., the authors utilize the euclidean distance sum in order to to measure uncertainty instead of the naive maximum softmax probability. The bounding boxes of the remaining proposal regions of unknown objects are further refined by an unsupervised clustering algorithm. 

While the aforementioned anomaly detection approaches excel in their individual modality setting, they oppose the great potential for improvement by combining the several modalities. Multimodal anomaly detection builds on the idea of complementarity. The individual modalities' advantages compensate for the weaknesses of the others to end up with a more reliable detection of anomalies. The idea of multimodal anomaly detection is adopted to anomaly detection by Wang~et~al.~\cite{wangRadarGhostTarget2021}. The authors use multimodal transformers to detect "ghost targets", i.e. radar reflections of vehicles, based on the affinity of lidar and radar point cloud data. But the technique is limited to the detection of the special case of multipath reflections. Other line of work base their multimodal anomaly detection on RGB-D data \cite{sunRealtimeFusionNetwork2020, guptaMergeNetDeepNet2018}. While there are camera and lidar fusion networks \cite{zhaoFusion3DLIDAR2020}, current literature misses anomaly detection across camera and lidar data. Furthermore, in contrast to recent works on fusion approaches such as~\cite{zhao_lidar-camera_2019}, we focus on open-set scenarios~\cite{wongIdentifyingUnknownInstances2019}. Hence, we propose a state-of-the-art road anomaly detection approach that propagates initial corner case proposals through the two sensor modalities and filters out known classes in 3D lidar as well as 2D camera space.

\section{Method}
\label{sec:method}


\begin{figure*}[h!]
\centering
\resizebox{\textwidth}{!}{
\includegraphics{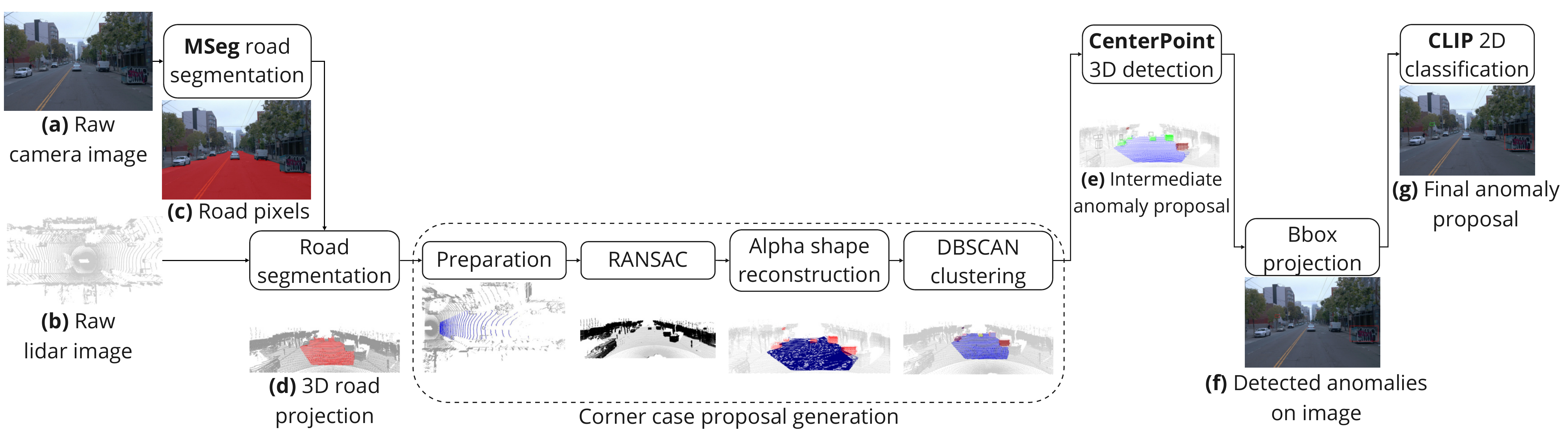}}
\caption{The flowchart gives an overview of our anomaly detection pipeline. The main components are explained in more detail in the Sections~\ref{subsec:Road_seg}-\ref{subsec:camera_det}. For each scene, we make use of camera \textbf{(a)} and lidar \textbf{(b)} data. Semantic segmentation is applied to the camera image to identify the road, our area of interest \textbf{(c)}. These road pixels are then projected into the lidar space~ \textbf{(d)}. The following corner case proposal generation is divided into several steps. First, the point cloud is cropped to only consider the front view. Next, the projected road plane is re-estimated using \ac{RANSAC} to account for inaccuracies. Alpha shape reconstruction of the plane is used to identify the points of objects that lay on the road. Those points are then clustered via \ac{DBSCAN} to generate corner case proposals. Next, CenterPoint object detection is applied, and cluster without classifications are marked as anomalies, creating an intermediate anomaly proposal \textbf{(e)}. The bounding boxes of those possible anomalies are then mapped onto the 2D camera space \textbf{(f)}. Based on a list of classes we deem normal, CLIP classifies them as either an anomaly or one of the non-anomaly classes, producing the final anomaly detection \textbf{(g)}. 
}
\label{fig:pipeline-overview}
\end{figure*}

We suggest a novel pipeline for detecting unknown objects by using lidar and camera data. In order to focus on the relevant driving area, we apply semantic segmentation on the input image. As a result, we derive a mask of the road coordinates, which we map into the 3D lidar space. We perform clustering, which we define as the source of truth that some type of object is present. These are then compared with detections from a 3D object detector. The points of clustered objects that could not be detected are transformed into the 2D image space, which has richer information. Here, we apply an image classifier. Only if the image classifier is not able to classify the object, we define it as an anomaly. Fig.~\ref{fig:pipeline-overview}
provides an overview of our anomaly detection pipeline.

\subsection{Road Segmentation}
\label{subsec:Road_seg}
In order to focus primarily on objects on the road, we applied a semantic segmentation model in the image domain as a first step. By using camera data, we made use of the higher information density through given shape and texture properties compared to the sparse lidar point cloud\cite{Survey_3D_object_detection}. As the Waymo Open Dataset does not provide pixel-wise labeled data, no semantic segmentation model is trained and tested on this dataset. Therefore, we used a pre-trained model based on the MSeg dataset~\cite{Lambert2020MSegAC}. MSeg is a  dataset that combines datasets with semantic segmentation groundtruth from multiple domains, such as COCO~~\cite{Lin2014MicrosoftCC}, ADE20K\cite{Zhou2018SemanticUO}, Mapillary~\cite{Mapillary}, IDD~\cite{Varma2019IDDAD}, BDD100K~\cite{Yu2020BDD100KAD}, Cityscapes~\cite{Cityscapes}, and SUN RGB-D~\cite{SUN_RGBD}. The model uses the HRNet-W48\cite{Sun2019HighResolutionRF} architecture and results in a higher generalization to unseen datasets. It is ranked first on the WildDash~\cite{Wilddash} leaderboard benchmark, which allows "for testing the robustness of models trained on other datasets". The model reaches an mIoU of 63.5 on BDD100K and 76.3 on Cityscapes. After obtaining the road pixels of each image, we transform these into the 3D lidar space. This allows us to receive a road mask to determine which objects are of interest. 

\subsection{Corner case proposal generation}
As we solely focus on anomalies on the road, we further processed the 3D road projection and raw lidar point cloud. First of all, we cropped the 360°-lidar data to a front view and excluded all information behind the ego-vehicle's center. While MSeg did show high performance and generalization to unseen datasets, we further processed the road segmentation. The transition between road segmentation and objects like cars or other traffic participants is not always a clear-cut. We fixed this inaccuracy to some degree by re-estimating the plane of the projected road segmentation via \ac{RANSAC}\cite{fischlerRandomSampleConsensus1981}. We sampled a random plane 500~times based on ten points. We then considered the remaining points as outliers of the sampled plane, whenever they exceeded the distance of 0.5~m. Thereby, we filtered out most points that stick out of the road plane in the direction of the z-axis due to inaccuracies of the road segmentation at the bottom of objects. Besides the re-estimation of the road plane, we also removed statistical outliers which would otherwise deteriorate the subsequent road surface estimation in the identified road points. Therefore, we calculated the average and standard deviation of the distances between points based on the 20 nearest neighbors. As lidar data become sparser with distance from the lidar sensor origin, we chose a considerably high standard deviation ratio of eight to exclude points. This means road points were considered as outliers whenever the standard deviation exceeded eight times the average standard deviation of all points' distances to their neighbors.

In order to determine whether objects lay on the road or not, we reconstructed the road surface based on the cleaned road points. However, the estimation of concave hulls occupied by a set of points is challenging. A simple and computationally efficient algorithm for surface reconstruction is the alpha shape estimation by Edelsbrunner~et~al.~\cite{edelsbrunnerShapeSetPoints1983}. The parameter $\alpha$ controls for the tightness of the polygon around the given road points, where an $\alpha$ of zero corresponds to the set of points as it is and an $\alpha$ towards infinity is the convex hull. As lidar data are sparse and leave the space empty behind projected objects, we found an $\alpha=10$ as a good trade-off value between a gap-less, crude, and a fine reconstruction of the road. 
Next, we set up the estimated road surface as a ray casting scene and check whether the non-road points are standing vertically on the road. Therefore, we flattened all points and reduced the point cloud to two dimensions, i.e., setting the height to zero. A point lies on the road if the number of intersections with the scene of the estimated alpha shape is even, and thus the flattened point lies within the surface.

After we identified the point cloud on the road, we generated corner case proposals by clustering the points via \ac{DBSCAN}\cite{esterDensityBasedAlgorithmDiscovering}. We chose an $\epsilon=1$ as a distance to neighbors in a cluster and required that a cluster consists of at least 30 points. The clustering into single objects makes our pipeline an open-set detection, as clusters are initially considered unknown. However, we used state-of-the-art closed-set detection architectures to rule out clusters as known objects.

\subsection{3D Lidar Detection}
\label{subsec:lidar_det}
We first utilized CenterPoint++~\cite{yinCenterbased3DObject2021} to label objects as known in the 3D lidar space. CenterPoint++ ranked 2nd in the Waymo Real-time 3D detection challenge~\cite{sunScalabilityPerceptionAutonomous2020}, as it achieved an mAPH of 72.8 and an inference speed of 57.1ms. Besides the outstanding performance, we chose CenterPoint++ as the model is only based on lidar data and its implementation is open-sourced. To be more precise, we used the two-staged model architecture with the VoxelNet~\cite{zhouVoxelNetEndtoEndLearning2018} backbone. Moreover, the model's input is the multi-sweep aggregation of the current and the last two-point cloud frames. Finally, we labeled an object cluster as known whenever at least half of the points fell into the bounding box detection of CenterPoint++. The remaining, uncovered objects are marked as anomalies in 3D lidar data. 

\subsection{2D camera detection}
\label{subsec:camera_det}
Clustered objects that were not classified by the 3D detector are further processed in the 2D image space. Therefore, we mapped the detected 3D bounding boxes onto the corresponding position in the image space. Next, we constructed 2D candidate bounding boxes from the area of each 3D box and passed them to the 2D classifier for recognition. To decide whether an object is classified as an anomaly, we used a simple threshold of the softmax classifier probability of $0.25$.  

For the 2D classification, we used CLIP~\cite{radford_learning_2021}, a zero-shot model that can be used for visual classification tasks by providing the names of the visual categories to be recognized.
CLIP learns visual concepts from natural language supervision by training an image encoder and a text encoder to predict, for a given batch of image-text pairings, which images were actually paired with which text snippets in the dataset. To do so, CLIP trains the image and text encoders to maximize the cosine similarity of the embeddings of the correct image-text pairs and to minimize the similarity of the embeddings of the incorrect pairs. A ResNet-50~\cite{he_deep_2016} and Vision Transformer (ViT)~\cite{dosovitskiy_image_2021} are used for encoding the images, and a transformer model for text encoding. CLIP is trained on $400$ million image-text pairs gathered from the internet. To use CLIP for zero-shot classification, the feature embeddings of the image and the potential text pairs, which are the classes of the dataset, are computed by the encoders. Next, the cosine similarity is calculated to find the closest image-text pair and softmax normalization is applied to get the probability distribution. When providing the labels, engineering a prompt such as \textit{"A photo of a {label}"} and customizing it to the given task has been shown to improve performance over using just the label. CLIP's zero-shot capability is competitive to fully supervised models on several benchmarks and tasks without pre-training on a specific dataset. For example, on ImageNet~\cite{russakovsky_imagenet_2015} zero-shot, CLIP matches the accuracy of a pre-trained ResNet-50. Since CLIP needs no pre-training for convincing zero-shot accuracy and the labels can freely be chosen, we are not restricted to a set of predetermined object categories. This fits our aim of anomaly detection, since we can define classes of objects that commonly appear in a driving setting as our non-anomaly classes and are not restricted to labels of datasets like Waymo, where only \textit{vehicles, pedestrians, cyclists}, and \textit{signs} are used. We used the classes with the highest frequency in the Waymo dataset as known classes. For that, we applied the Detr~\cite{carion_end--end_2020} 2D detector, pre-trained on the COCO~\cite{lin_microsoft_2014} dataset, to detect objects in the 2D images. We took the classes corresponding to the top $99\%$ of the detected objects and ended up with the following $13$ classes as the classes not considered anomalies for our approach: car, traffic light, person, truck, bus, fire hydrant, bicycle, handbag, backpack, parking meter, stop sign, umbrella, and motorcycle. We further added the classes tree, pole, and bush, since these objects appear frequently in the Waymo dataset, but are not part of the COCO dataset labels. This list can be dynamically extended as needed. One interesting direction might be the inclusion of Cityscape classes~\cite{Cityscapes}, since these are often used in research for detection approaches.

We used the prompt \textit{"A photo of a {label} on a street“} which is more adapted to our task.

Furthermore, CLIP is shown to be much more robust than supervised ImageNet models with equivalent accuracy. \ac{dl} models trained on a specific dataset often learn correlations of the specific data distribution they were trained on to improve in-distribution performance. However, this dataset specialized training often leads to worse performance on new dataset. Since zero-shot models like CLIP cannot exploit such correlations, they are more robust to distribution shifts. In our case where objects are captured under challenging conditions in which the scale of the object varies heavily, CLIP still performs due to its generalization ability and robustness.

Using a threshold of the classifier's prediction probability to identify an anomaly can be problematic, since it has been shown that the prediction probability from a softmax distribution does not directly correspond to the confidence of the model~\cite{guo_calibration_2017, nguyen_deep_2015}. Only for calibrated models, the produced prediction score can directly be interpreted
as the confidence of the model. However,~\cite{hendrycks_baseline_2017} discovered that the prediction probability of out-of-distribution examples, in our case the anomalies, is, in general, lower than for in-distribution examples, the non-anomalies. Furthermore, CLIP has been proven to be well-calibrated for in-and out-of-distribution datasets~\cite{minderer_revisiting_2021}. Therefore, this threshold method can be used as a valid baseline approach for anomaly detection. 

\section{Evaluation}
\label{sec:evaluation}

\begin{figure*}[ht!]
\setlength\tabcolsep{3pt} 
\centering
\begin{tabular}{@{} r M{0.185\linewidth} M{0.185\linewidth} M{0.185\linewidth} M{0.185\linewidth} @{}}
& MSeg & Clustering & CenterPoint & CLIP\\
    \begin{subfigure}{0.05\linewidth}\caption{}
    \label{subfig:a} 
    \end{subfigure} 
    & \includegraphics[width=\hsize]{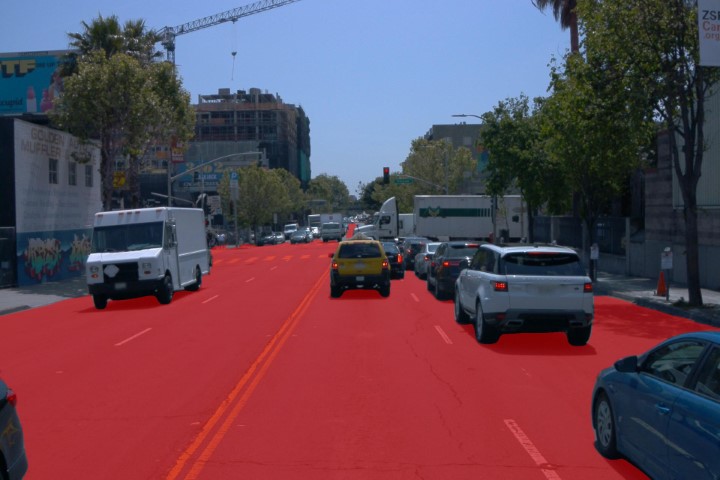} 
    & \includegraphics[width=\hsize]{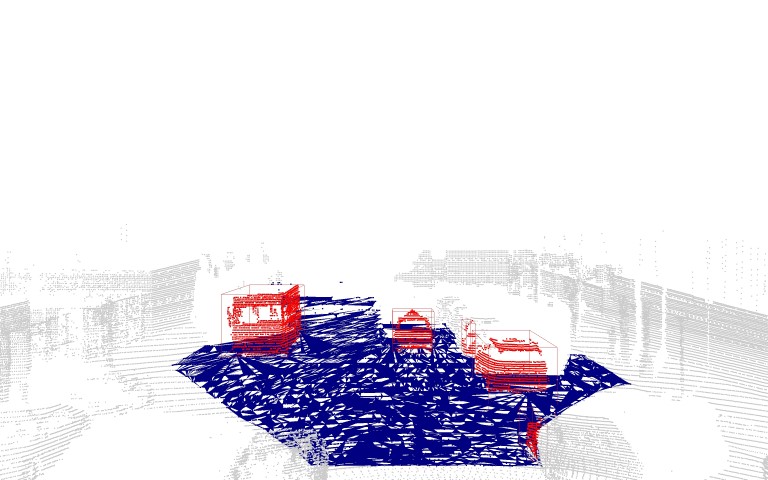}
    & \includegraphics[width=\hsize]{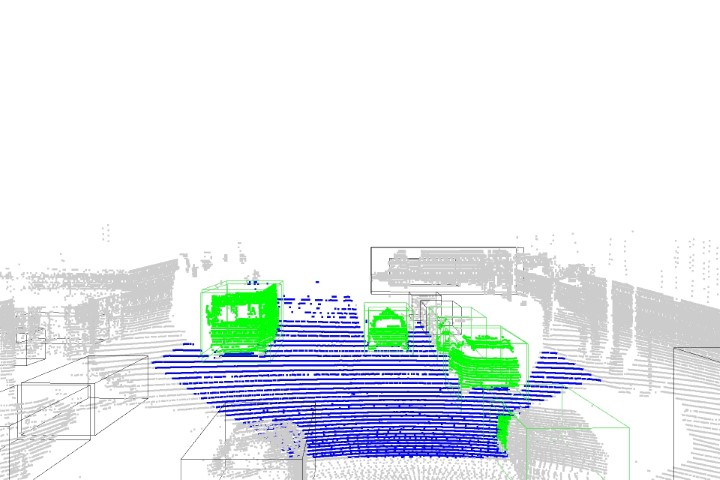}    
    & \includegraphics[width=\hsize]{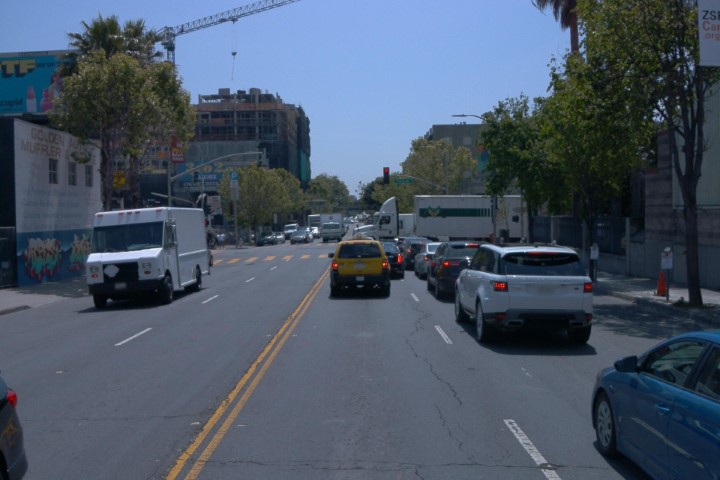}\\ 
    \addlinespace
    \begin{subfigure}{0.05\linewidth}\caption{}
    \label{subfig:b} 
    \end{subfigure} 
    & \includegraphics[width=\hsize]{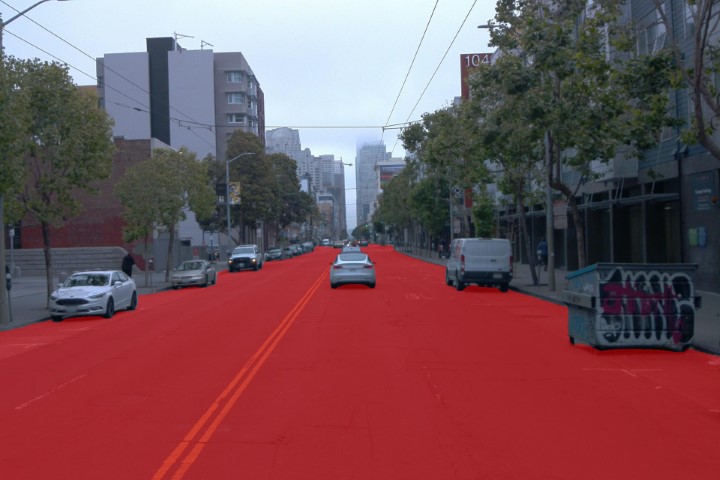} 
    & \includegraphics[width=\hsize]{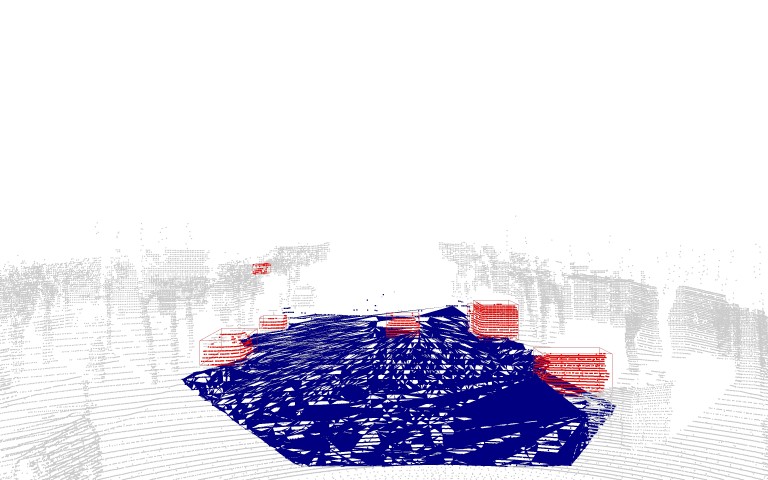}
    & \includegraphics[width=\hsize]{img/lidar/7240042450405902042_580_000_600_000-1559312909237488.jpg}    
    & \includegraphics[width=\hsize]{img/camera/7240042450405902042_580_000_600_000-1559312909237488.jpg}\\ 
    \addlinespace
    \begin{subfigure}{0.05\linewidth}\caption{}
    \label{subfig:c} 
    \end{subfigure} 
    & \includegraphics[width=\hsize]{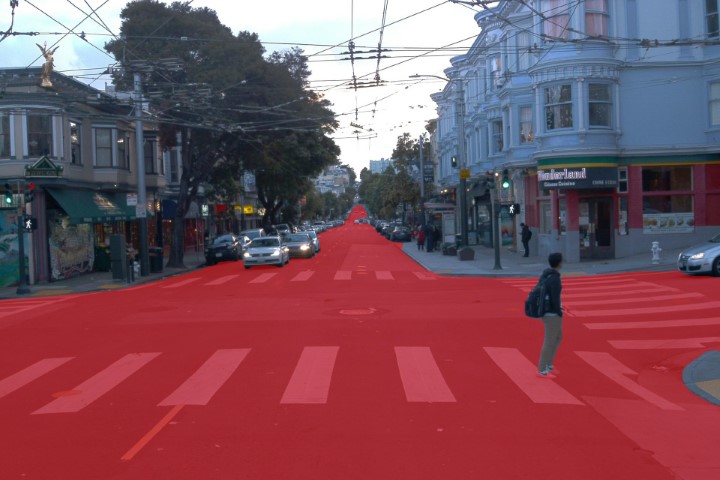} 
    & \includegraphics[width=\hsize]{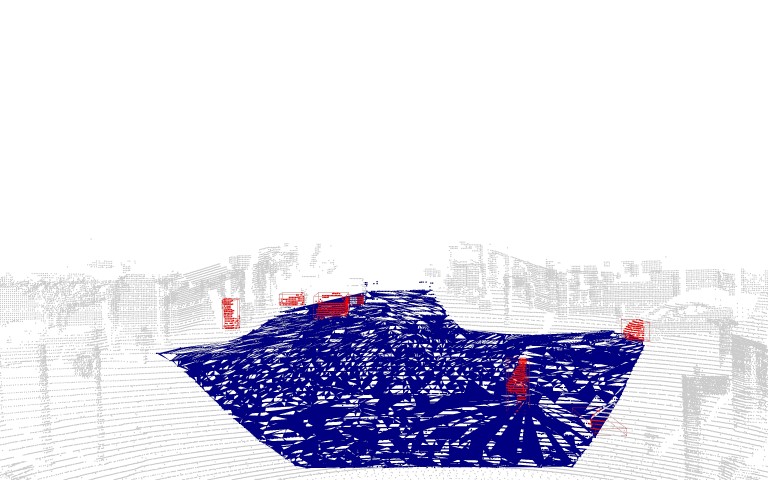}
    & \includegraphics[width=\hsize]{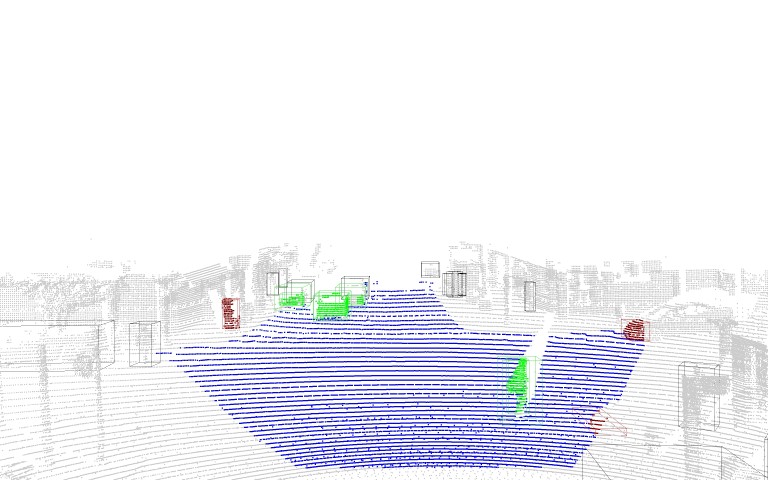}    
    & \includegraphics[width=\hsize]{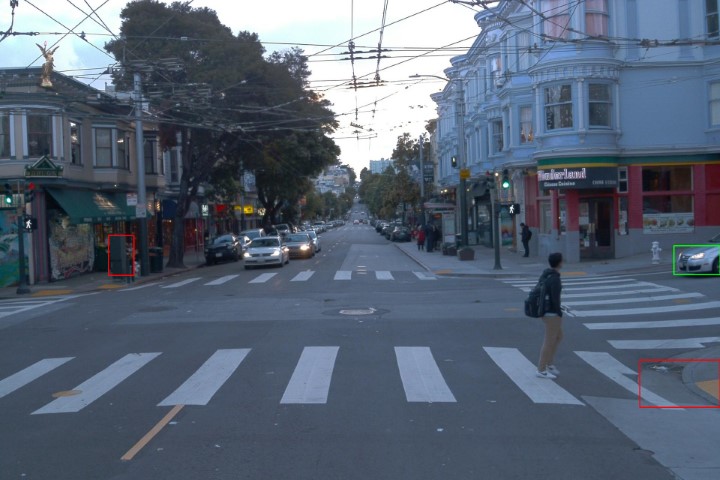}\\ 
    \addlinespace
    \begin{subfigure}{0.05\linewidth}\caption{}
    \label{subfig:d} 
    \end{subfigure} 
    & \includegraphics[width=\hsize]{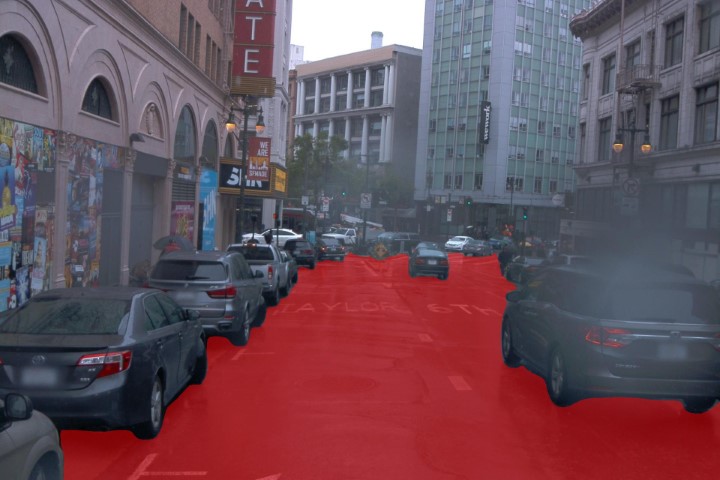} 
    & \includegraphics[width=\hsize]{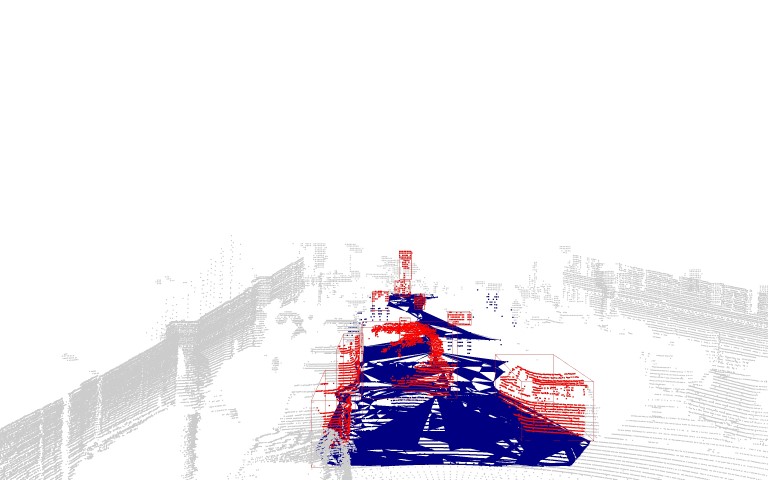}
    & \includegraphics[width=\hsize]{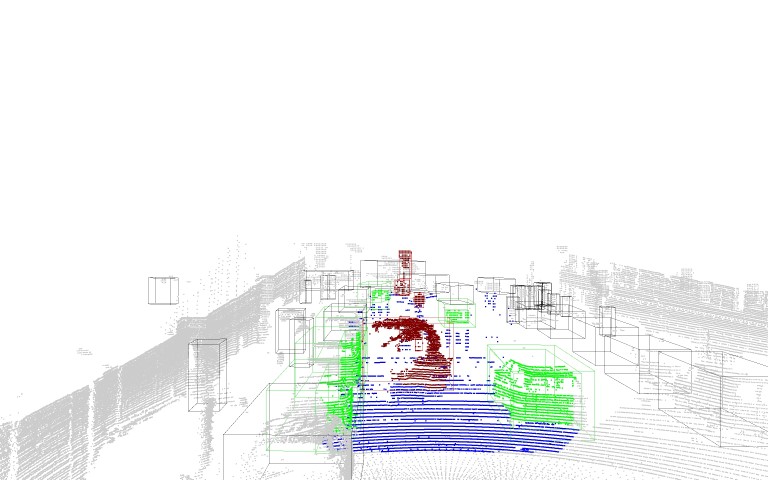}    
    & \includegraphics[width=\hsize]{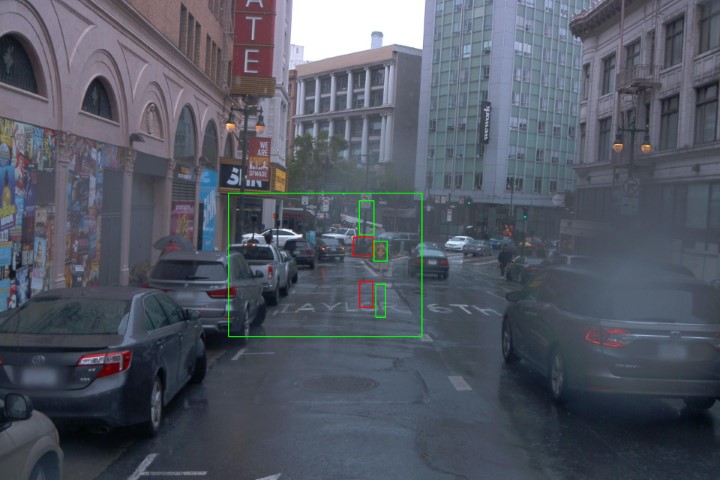}\\ 
    \addlinespace 
    \begin{subfigure}{0.05\linewidth}\caption{}
    \label{subfig:e} 
    \end{subfigure} 
    & \includegraphics[width=\hsize]{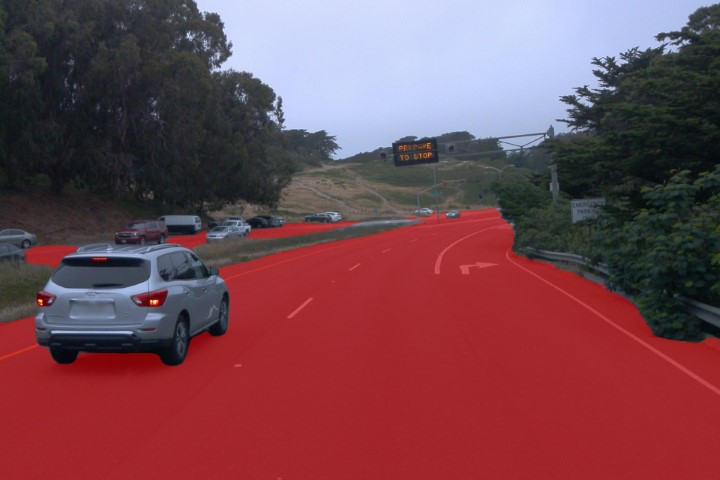} 
    & \includegraphics[width=\hsize]{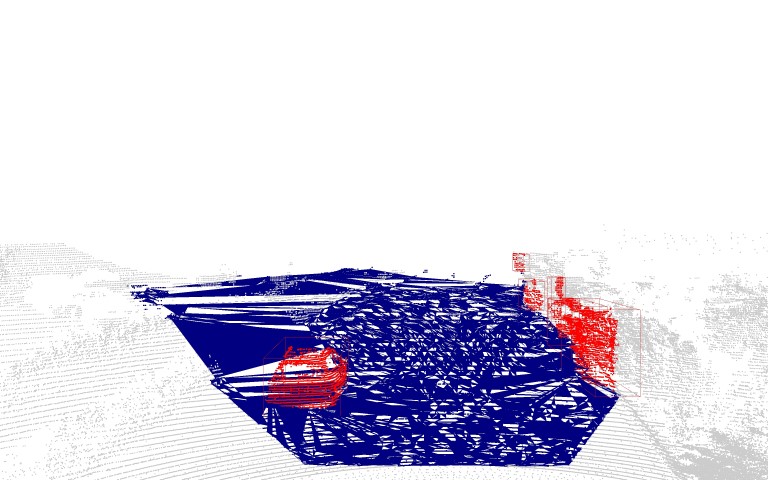}
    & \includegraphics[width=\hsize]{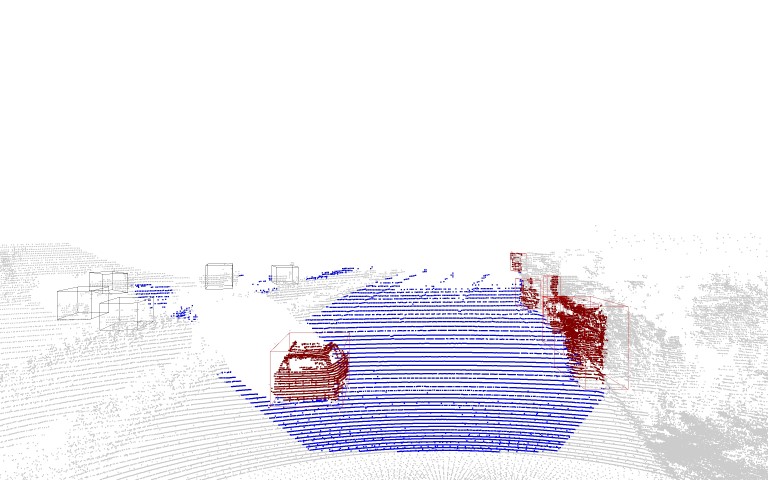}    
    & \includegraphics[width=\hsize]{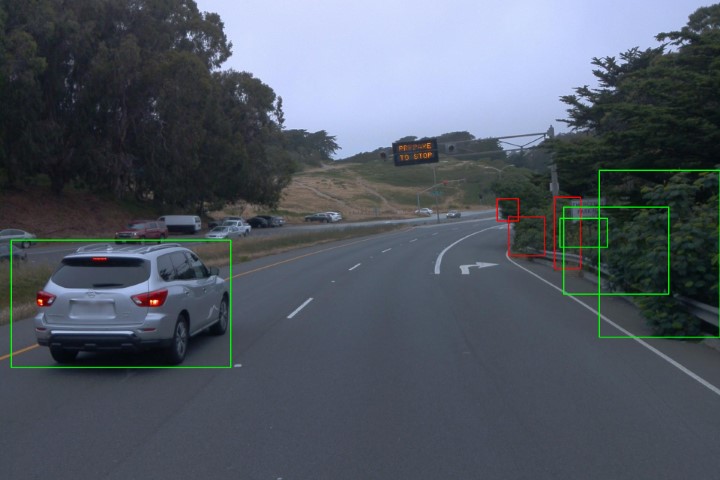}\\ 
    \addlinespace
    \begin{subfigure}{0.05\linewidth}\caption{}
    \label{subfig:f} 
    \end{subfigure} 
    & \includegraphics[width=\hsize]{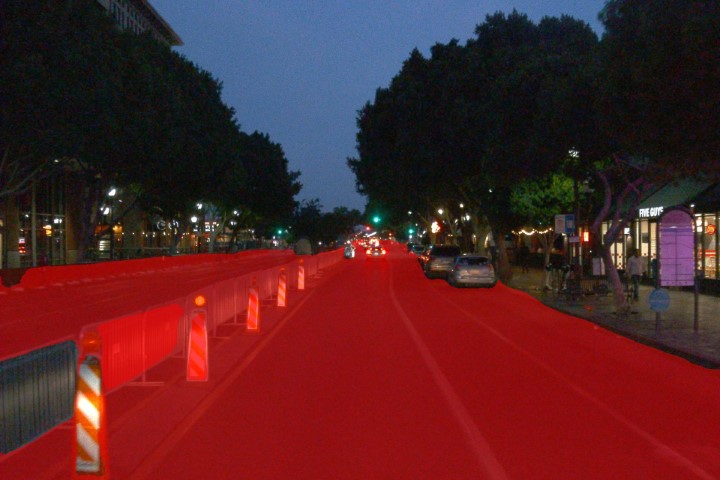} 
    & \includegraphics[width=\hsize]{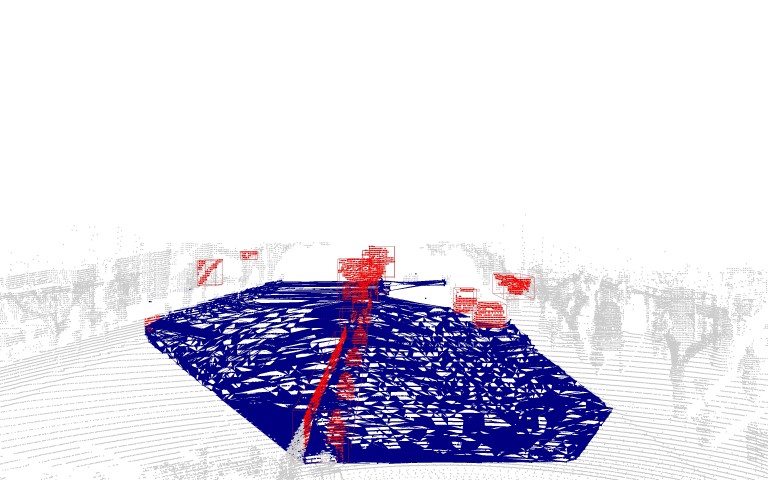}
    & \includegraphics[width=\hsize]{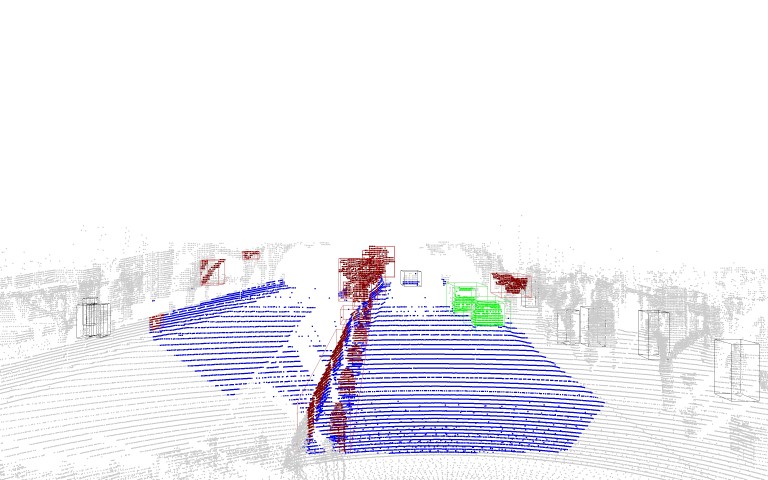}    
    & \includegraphics[width=\hsize]{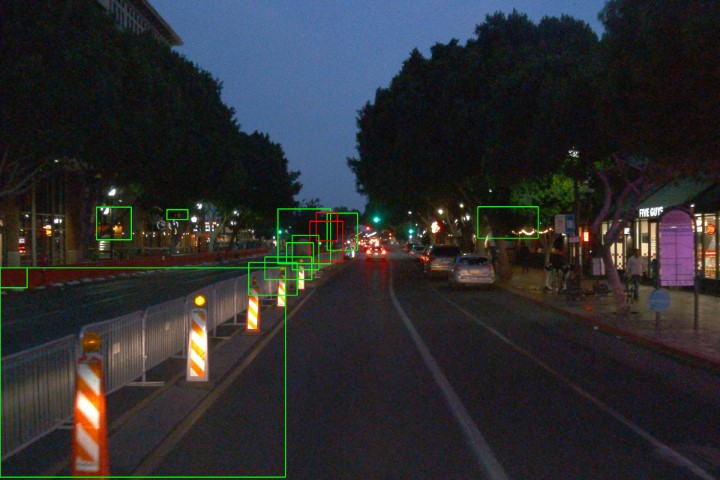}\\ 
    \addlinespace
    \begin{subfigure}{0.05\linewidth}\caption{}
    \label{subfig:g} 
    \end{subfigure} 
    & \includegraphics[width=\hsize]{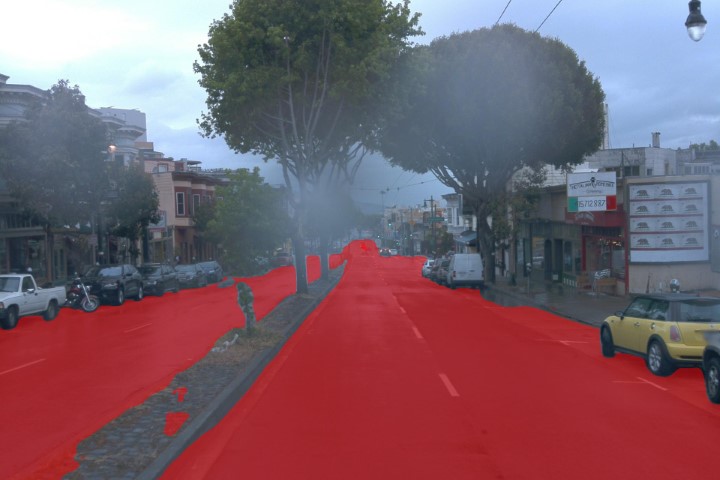} 
    & \includegraphics[width=\hsize]{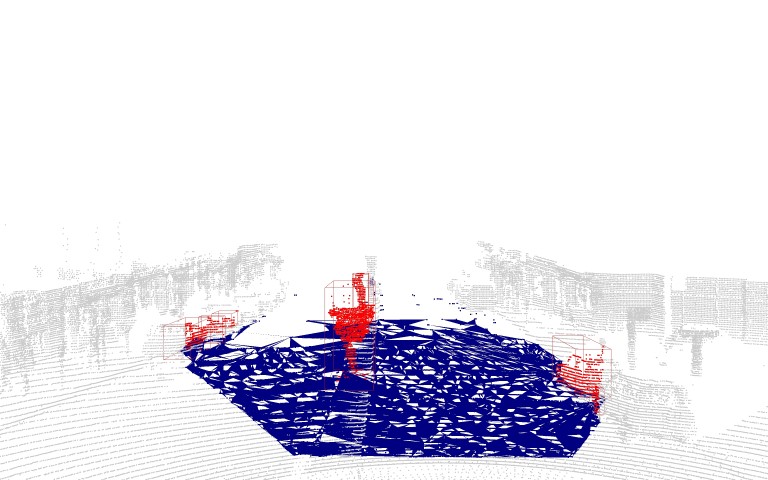}
    & \includegraphics[width=\hsize]{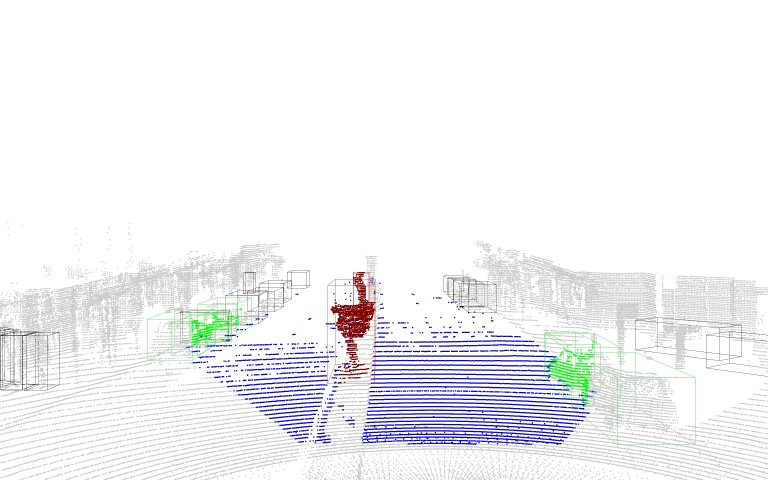}    
    & \includegraphics[width=\hsize]{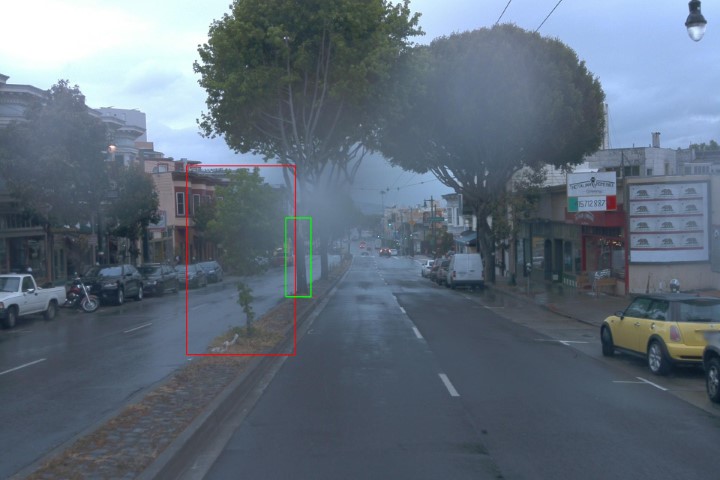}\\ 
    \addlinespace
    \begin{subfigure}{0.05\linewidth}\caption{}
    \label{subfig:h} 
    \end{subfigure} 
    & \includegraphics[width=\hsize]{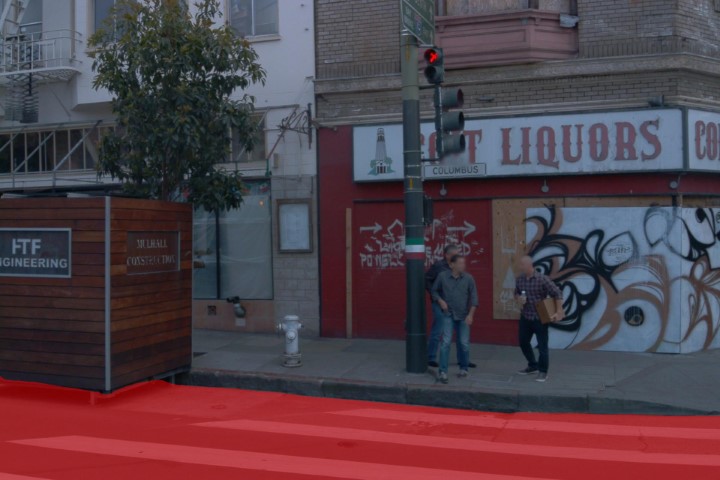} 
    & \includegraphics[width=\hsize]{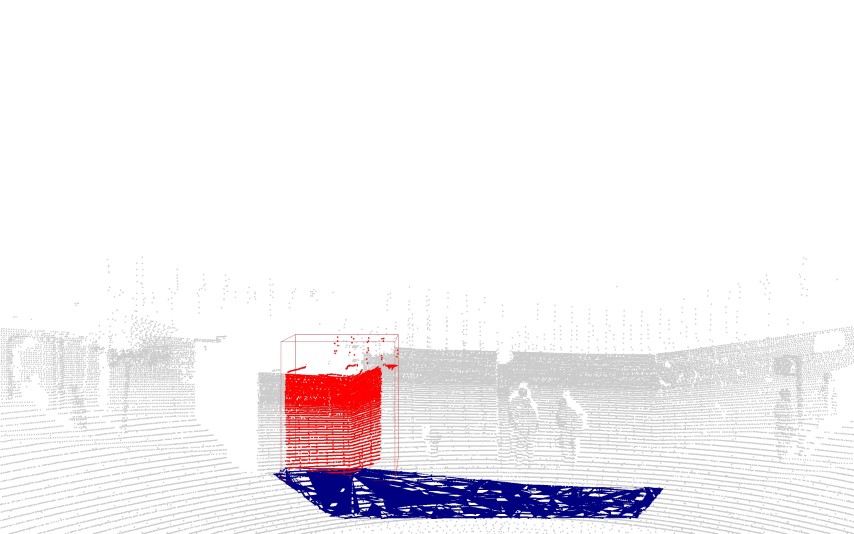}
    & \includegraphics[width=\hsize]{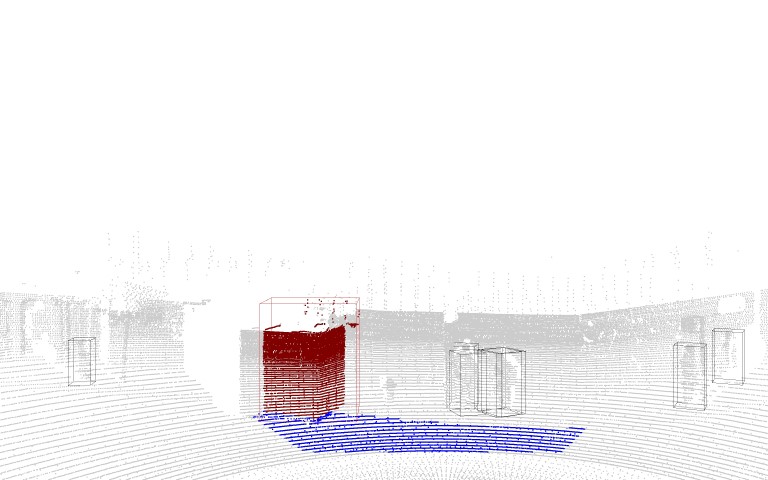}    
    & \includegraphics[width=\hsize]{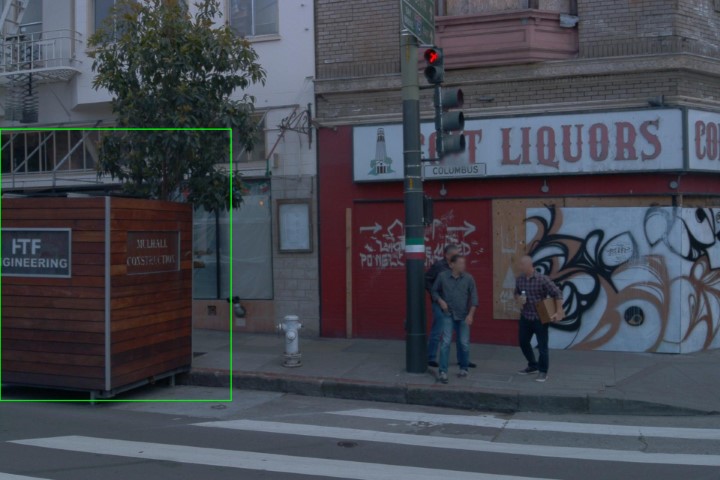}\\ 
    \addlinespace
\end{tabular}
\caption{Exemplary results from our pipeline: The Subfigures \ref{subfig:a}-\ref{subfig:h} correspond to different types of detection cases or failures. The four labeled columns visualize the intermediate results of the four pipeline components: Road segmentation (MSeg), corner case proposal generation (clustering), 3D lidar detection (CenterPoint), and 2D camera detection (CLIP). The first column shows the road segmentation of MSeg in red. The clustering column visualizes the alpha shape estimation in blue and the clustered point cloud on the road (red). Next, the clusters covered by CenterPoint are colored in green. The last column outlines final anomalies in red bounding boxes while representing lidar anomalies covered by CLIP in green.}
\label{fig:examples}
\end{figure*}

Due to the lack of ground truth labels for anomalies in public datasets so far, we cannot provide a quantitative evaluation of our method and are restricted to qualitative evaluation only. 
Our method faced several challenges, we therefore define different typical scenarios that occurred and present the respective output of each of our pipeline steps, see Fig.~\ref{fig:examples}. 
\paragraph{All objects detected in lidar}
The most common scenario is when all clustered objects are already detected by CenterPoint and there is no possible anomaly that the 2D classification could detect, Fig.~\ref{subfig:a} provides an example.
\paragraph{True anomalies found on the road}
In the second category, our approach detects true anomalies on the road, as shown in Fig.~\ref{subfig:b}. MSeg correctly identified the road shown by the red area. The following step clusters the objects present in the scene, however in the top left of the frame, also a part of a tree is clustered. CenterPoint classified most of the objects, except for the part of the tree and the dumpster in the front. Since dumpsters are not part of the CLIP labels and are thus anomalies, they should be classified as an anomaly. CLIP indeed labels it as an anomaly, indicated by the red rectangle around the object.
\paragraph{True anomalies found off the road}
Fig.~\ref{subfig:c} shows a scenario where an anomaly on the left side of the image, probably a parking ticket machine, was correctly identified. However, since the object is on the sidewalk and not on the road, it should not have been clustered. To overcome this issue, the alpha shape estimation should be further refined. This scene also shows how a part of the sidewalk (bottom right) is clustered and falsely identified as an anomaly by CLIP, since it does not match with any of the known classes. 
\paragraph{Anomaly in lidar, but not in camera space}
Fig.~\ref{subfig:d} is an example where an anomaly is found in the 3D lidar space but not in the 2D camera space. In this case CenterPoint detected the smoke in front of the vehicle and identified is as an anomaly. However, CLIP did not recognize the smoke as an anomaly since it is not visible in the camera image. This also demonstrates the complementary capability of our pipeline and the advantages of using the two modalities.
\paragraph{Failure due to over-clustering}
A typically problematic scenario is over-clustering in the lidar space. The parts of the tree on the image in Fig.~\ref{subfig:e} that are reaching over the road are not clustered as one tree object, but as several smaller ones. CLIP then has problems classifying the objects with only a small part of it, and thus identifies some parts of the tree as anomalies in this case. Further, adjusting \ac{DBSCAN} could improve the cluster performance.   
\paragraph{Failure due to under-clustering}
The opposite of the over-clustering problem is when several objects are clustered as one and not individually, for example in Fig.~\ref{subfig:f}, where several street cones and the fence are seen as one object. 
\paragraph{Non-anomaly falsely identified as anomaly}
Besides true positive (TP) examples, there are also cases where a non-anomaly is classified as an anomaly. In this case, ~\ref{subfig:g} a tree, producing a false positive (FP). This scene also shows a problem with the alpha shape estimation. MSeg correctly identified that there is a separation of the road in the middle and thus does not label this part as road, but alpha shape takes the whole ground as the area of interest and therefore the tree is also clustered, even though it is not part of the road. 
\paragraph{True anomaly not identified}
The pipeline also produces false negatives (FN), i.e., objects that are anomalies are not identified as such. Fig.~\ref{subfig:h} shows how some kind of big box was correctly classified as anomaly in the lidar space but not recognized as such by the 2d classifier. Instead CLIP classified it as a truck.

The different scenarios show that our pipeline is able to make use of the different modalities and combines them to correctly identify anomalies. However, it also suffers from limitations of the models used in the respective steps, which can be addressed in future work. There are cases where the road segmentation is inaccurate and the area of interest is larger than the actual road. We have tried to mitigate this aspect with a road lane detection approach, which did not perform well in complex urban settings though. The alpha shape reconstruction of the road surface in lidar space is not always accurate, since parts of the sidewalk or the separation areas between the lines are taken into account. To improve the alpha shape estimation, one could fine tune the corresponding hyperparameter $\alpha$ on a driving dataset suited for semantic segmentation or consider multiple road surface within one scene. Alternatively, this issue can completely be mitigated with the utilization of map data. Furthermore, over-and under-clustering of \ac{DBSCAN} is problematic for the following 3D detection and 2D classification. Single objects are either not clustered individually, resulting in an area with several objects, or they are only clustered partly, making it difficult to identify the object. Similarly to the road surface estimation, one could further improve the clustering by fine-tuning the parameters on a different dataset or replace \ac{DBSCAN} by a more complex algorithm \cite{liInsClusteringInstantlyClustering2020}. When CenterPoint is unable to detect all objects, in many cases CLIP can provide an accurate classification, showing the complementary nature of the pipeline. However, our 2D classification step suffers from two limitations. Firstly, even though the CLIP model has a good zero-shot performance and is robust to distribution shifts, there are misclassifications due to the difficult conditions, such as varying scale, the mentioned over-and under-clustering, and environmental disturbances. Furthermore, the model sometimes provides inconsistent results, changing the classification output for objects of different frames of the same scenario. A more specialized prompt engineering could increase CLIP's performance. The second limitation is that we use a method based on a confidence threshold for identifying an anomaly. Although CLIP is well-calibrated, this approach should be seen as a baseline method and can be outperformed by more sophisticated anomaly detection methods, see Sec.~\ref{sec:related_work}).

Overall, the results show a strong potential of the proposed approach, as the semantic knowledge embedded in camera data can greatly benefit the lidar based anomaly proposals.

\section{Conclusion}
\label{sec:conclusion}

Our presented anomaly detection pipeline successfully combines camera and lidar modalities, utilizing their respective advantages. We focused on the aspect of combining different modalities and not on the model used for each domain. As mentioned, our threshold based anomaly detection approach can be seen as a strong baseline.

One major problem we faced during the development is the lack of publicly available multimodal datasets for anomaly detection. Without ground truth labels for anomalies in the dataset, a quantitative evaluation is not possible, and we could thus only provide an overview of typical scenarios we encountered. Just recently, Li et al. introduced CODA~\cite{li_coda_2022}, a real-world road corner case dataset consisting of different scenes from major real-world object detection benchmarks~\cite{geiger_are_2012, caesar_nuscenes_2020, mao_one_2021}, containing at least one corner case that poses a safety risk to self-driving vehicles or their surroundings. For each scene, there is the camera and corresponding lidar data provided. However, as of now, the full dataset has not yet been released. Future work can use this dataset for quantitative evaluation of anomaly detection.

\section{Acknowledgment}
\label{sec:acknowledgment}

This work results partly from the KIGLIS project supported by the German Federal Ministry of Education and Research (BMBF), grant number 16KIS1231.


{\small
\bibliographystyle{IEEEtran}
\bibliography{references}

\begin{thebibliography}{10}
\providecommand{\url}[1]{#1}
\csname url@samestyle\endcsname
\providecommand{\newblock}{\relax}
\providecommand{\bibinfo}[2]{#2}
\providecommand{\BIBentrySTDinterwordspacing}{\spaceskip=0pt\relax}
\providecommand{\BIBentryALTinterwordstretchfactor}{4}
\providecommand{\BIBentryALTinterwordspacing}{\spaceskip=\fontdimen2\font plus
\BIBentryALTinterwordstretchfactor\fontdimen3\font minus
  \fontdimen4\font\relax}
\providecommand{\BIBforeignlanguage}[2]{{%
\expandafter\ifx\csname l@#1\endcsname\relax
\typeout{** WARNING: IEEEtran.bst: No hyphenation pattern has been}%
\typeout{** loaded for the language `#1'. Using the pattern for}%
\typeout{** the default language instead.}%
\else
\language=\csname l@#1\endcsname
\fi
#2}}
\providecommand{\BIBdecl}{\relax}
\BIBdecl

\bibitem{Daimler_Level3}
\BIBentryALTinterwordspacing
{Mercedes-Benz Group}, ``First internationally valid system approval for
  conditionally automated driving,'' 2021, accessed: 12.02.2022. [Online].
  Available:
  \url{https://group.mercedes-benz.com/innovation/product-innovation/autonomous-driving/system-approval-for-conditionally-automated-driving.html}
\BIBentrySTDinterwordspacing

\bibitem{sae_2018}
{On-Road Automated Driving (ORAD) committee}, \emph{SAE-J3016: Taxonomy and
  Definitions for Terms Related to Driving Automation Systems for On-Road Motor
  Vehicles}, 2021.

\bibitem{Torc_L4}
\BIBentryALTinterwordspacing
{Torc Robotics}, ``Torc robotics to expand self-driving truck testing to new
  mexico with test center in albuquerque,'' 2020, accessed: 12.02.2022.
  [Online]. Available:
  \url{https://torc.ai/torc-robotics-to-expand-self-driving-truck-testing-to-new-mexico-with-test-center-in-albuquerque/}
\BIBentrySTDinterwordspacing

\bibitem{Tusimple_L4}
\BIBentryALTinterwordspacing
{Rebecca Bellan}, ``Tusimple completes its first driverless autonomous truck
  run on public roads,'' 2021, accessed: 12.02.2022. [Online]. Available:
  \url{https://techcrunch.com/2021/12/29/tusimple-completes-its-first-driverless-autonomous-truck-run-on-public-roads/}
\BIBentrySTDinterwordspacing

\bibitem{Waymo_L4}
\BIBentryALTinterwordspacing
{Waymo}, ``Expanding our testing in san francisco,'' 2021, accessed:
  12.02.2022. [Online]. Available:
  \url{https://blog.waymo.com/2021/02/expanding-our-testing-in-san-francisco.html}
\BIBentrySTDinterwordspacing

\bibitem{dibiasePixelwiseAnomalyDetection2021a}
G.~Di~Biase, H.~Blum, R.~Siegwart, and C.~Cadena, ``Pixel-wise {{Anomaly
  Detection}} in {{Complex Driving Scenes}},'' in \emph{{{IEEE}}/{{CVF
  Conference}} on {{Computer Vision}} and {{Pattern Recognition}}}, 2021.

\bibitem{Joseph2021TowardsOW}
K.~J. Joseph, S.~H. Khan, F.~S. Khan, and V.~N. Balasubramanian, ``Towards open
  world object detection,'' \emph{2021 IEEE/CVF Conference on Computer Vision
  and Pattern Recognition (CVPR)}, 2021.

\bibitem{Bogdoll_Description_2021_ICCV}
D.~Bogdoll, J.~Breitenstein, F.~Heidecker, M.~Bieshaar, B.~Sick,
  T.~Fingscheidt, and M.~Z\"{o}llner, ``{Description of Corner Cases in
  Automated Driving: Goals and Challenges},'' in \emph{Proceedings of the
  IEEE/CVF International Conference on Computer Vision (ICCV) Workshops}, 2021.

\bibitem{wongIdentifyingUnknownInstances2019}
K.~Wong, S.~Wang, M.~Ren, M.~Liang, and R.~Urtasun, ``Identifying {{Unknown
  Instances}} for {{Autonomous Driving}},'' \emph{arXiv:1910.11296}, 2019.

\bibitem{sunScalabilityPerceptionAutonomous2020}
P.~Sun, H.~Kretzschmar, X.~Dotiwalla, A.~Chouard, V.~Patnaik, P.~Tsui, J.~Guo,
  Y.~Zhou, Y.~Chai, B.~Caine, V.~Vasudevan, W.~Han, J.~Ngiam, H.~Zhao,
  A.~Timofeev, S.~Ettinger, M.~Krivokon, A.~Gao, A.~Joshi, Y.~Zhang, J.~Shlens,
  Z.~Chen, and D.~Anguelov, ``Scalability in perception for autonomous driving:
  Waymo open dataset,'' in \emph{Proceedings of the IEEE/CVF Conference on
  Computer Vision and Pattern Recognition (CVPR)}, 2020.

\bibitem{lisDetectingUnexpectedImage2019}
K.~Lis, K.~K. Nakka, P.~Fua, and M.~Salzmann, ``Detecting the {{Unexpected}}
  via {{Image Resynthesis}},'' in \emph{{{IEEE}}/{{CVF International
  Conference}} on {{Computer Vision}}}, 2019.

\bibitem{nitschOutofDistributionDetectionAutomotive2021}
J.~Nitsch, M.~Itkina, R.~Senanayake, J.~Nieto, M.~Schmidt, R.~Siegwart, M.~J.
  Kochenderfer, and C.~Cadena, ``Out-of-{{Distribution Detection}} for
  {{Automotive Perception}},'' in \emph{{{IEEE International Intelligent
  Transportation Systems Conference}}}, 2021.

\bibitem{bogdollAnomalyDetectionAutonomous2022}
D.~Bogdoll, M.~Nitsche, and J.~M. Z{\"o}llner, ``Anomaly {{Detection}} in
  {{Autonomous Driving}}: {{A Survey}},'' \emph{arXiv:2204.07974}, 2022.

\bibitem{9304789}
J.~Breitenstein, J.-A. Termöhlen, D.~Lipinski, and T.~Fingscheidt,
  ``Systematization of corner cases for visual perception in automated
  driving,'' in \emph{2020 IEEE Intelligent Vehicles Symposium (IV)}, 2020.

\bibitem{Bogdoll_Addatasets_2022_VEHITS}
D.~Bogdoll, F.~Schreyer, and J.~M. Z\"{o}llner, ``{ad-datasets: a
  meta-collection of data sets for autonomous driving},''
  \emph{arXiv:2202.01909}, 2022.

\bibitem{sun2020scalability}
P.~Sun, H.~Kretzschmar, X.~Dotiwalla, A.~Chouard, V.~Patnaik, P.~Tsui, J.~Guo,
  Y.~Zhou, Y.~Chai, B.~Caine \emph{et~al.}, ``Scalability in perception for
  autonomous driving: Waymo open dataset,'' in \emph{Proceedings of the
  IEEE/CVF Conference on Computer Vision and Pattern Recognition}, 2020.

\bibitem{Bogdoll_Taxonomy_2022_FICC}
D.~Bogdoll, S.~Orf, L.~T\"{o}ttel, and J.~M. Z\"{o}llner, ``{Taxonomy and
  Survey on Remote Human Input Systems for Driving Automation Systems},'' in
  \emph{Advances in Information and Communication}.\hskip 1em plus 0.5em minus
  0.4em\relax Springer, 2022.

\bibitem{kendallBayesianSegNetModel2016}
A.~Kendall, V.~Badrinarayanan, and R.~Cipolla, ``Bayesian {{SegNet}}: {{Model
  Uncertainty}} in {{Deep Convolutional Encoder-Decoder Architectures}} for
  {{Scene Understanding}},'' \emph{arXiv:1511.02680}, 2016.

\bibitem{jungStandardizedMaxLogits2021a}
S.~Jung, J.~Lee, D.~Gwak, S.~Choi, and J.~Choo, ``Standardized max logits:
  {{A}} simple yet effective approach for identifying unexpected road obstacles
  in urban-scene segmentation,'' in \emph{{{IEEE}}/{{CVF}} International
  Conference on Computer Vision}, 2021.

\bibitem{heideckerCornerCaseDetection2021}
F.~Heidecker, A.~Hannan, M.~Bieshaar, and B.~Sick, ``Towards {{Corner Case
  Detection}} by {{Modeling}} the {{Uncertainty}} of {{Instance Segmentation
  Networks}},'' \emph{Pattern {{Recognition}}. {{ICPR International Workshops}}
  and {{Challenges}}}, vol. 12664, 2021.

\bibitem{huangEfficientUncertaintyEstimation2018}
P.-Y. Huang, W.-T. Hsu, C.-Y. Chiu, T.-F. Wu, and M.~Sun, ``Efficient
  {{Uncertainty Estimation}} for {{Semantic Segmentation}} in {{Videos}},''
  \emph{Computer {{Vision}} \textendash{} {{ECCV}}}, vol. 11205, 2018.

\bibitem{duVOSLearningWhat2022}
X.~Du, Z.~Wang, M.~Cai, and Y.~Li, ``{{VOS}}: {{Learning What You Don}}'t
  {{Know}} by {{Virtual Outlier Synthesis}},'' \emph{arXiv:2202.01197}, 2022.

\bibitem{chanEntropyMaximizationMeta2021a}
R.~Chan, M.~Rottmann, and H.~Gottschalk, ``Entropy maximization and meta
  classification for out-of-distribution detection in semantic segmentation,''
  in \emph{{{IEEE}}/{{CVF}} International Conference on Computer Vision}, 2021.

\bibitem{grcicDenseAnomalyDetection2021}
M.~Grci{\'c}, P.~Bevandi{\'c}, Z.~Kalafati{\'c}, and S.~{\v S}egvi{\'c},
  ``Dense anomaly detection by robust learning on synthetic negative data,''
  \emph{arXiv:2112.12833}, 2021.

\bibitem{ohgushiRoadObstacleDetection2021}
T.~Ohgushi, K.~Horiguchi, and M.~Yamanaka, ``Road {{Obstacle Detection Method
  Based}} on an {{Autoencoder}} with {{Semantic Segmentation}},''
  \emph{Computer {{Vision}} \textendash{}}, vol. 12627, 2021.

\bibitem{vojirRoadAnomalyDetection2021a}
T.~Vojir, T.~{\v S}ipka, R.~Aljundi, N.~Chumerin, D.~O. Reino, and J.~Matas,
  ``Road anomaly detection by partial image reconstruction with segmentation
  coupling,'' in \emph{{{IEEE}}/{{CVF}} International Conference on Computer
  Vision}, 2021.

\bibitem{leeSimpleUnifiedFramework2018}
K.~Lee, K.~Lee, H.~Lee, and J.~Shin, ``A simple unified framework for detecting
  out-of-distribution samples and adversarial attacks,'' \emph{Advances in
  Neural Information Processing Systems}, vol.~31, 2018.

\bibitem{blumFishyscapesBenchmarkMeasuring2021}
H.~Blum, P.-E. Sarlin, J.~Nieto, R.~Siegwart, and C.~Cadena, ``The
  {{Fishyscapes Benchmark}}: {{Measuring Blind Spots}} in {{Semantic
  Segmentation}},'' \emph{International Journal of Computer Vision}, vol. 129,
  2021.

\bibitem{cenOpenset3DObject2021}
J.~Cen, P.~Yun, J.~Cai, M.~Y. Wang, and M.~Liu, ``Open-set {{3D Object
  Detection}},'' \emph{arXiv:2112.01135}, 2021.

\bibitem{wangRadarGhostTarget2021}
L.~Wang, S.~Giebenhain, C.~Anklam, and B.~Goldluecke, ``Radar {{Ghost Target
  Detection}} via {{Multimodal Transformers}},'' \emph{IEEE Robotics and
  Automation Letters}, vol.~6, 2021.

\bibitem{sunRealtimeFusionNetwork2020}
L.~Sun, K.~Yang, X.~Hu, W.~Hu, and K.~Wang, ``Real-time {{Fusion Network}} for
  {{RGB-D Semantic Segmentation Incorporating Unexpected Obstacle Detection}}
  for {{Road-driving Images}},'' \emph{arXiv:2002.10570}, 2020.

\bibitem{guptaMergeNetDeepNet2018}
K.~Gupta, S.~A. Javed, V.~Gandhi, and K.~M. Krishna, ``{{MergeNet}}: {{A Deep
  Net Architecture}} for {{Small Obstacle Discovery}},'' in \emph{{{IEEE
  International Conference}} on {{Robotics}} and {{Automation}}}, 2018.

\bibitem{zhaoFusion3DLIDAR2020}
X.~Zhao, P.~Sun, Z.~Xu, H.~Min, and H.~Yu, ``Fusion of {{3D LIDAR}} and
  {{Camera Data}} for {{Object Detection}} in {{Autonomous Vehicle
  Applications}},'' \emph{IEEE Sensors Journal}, 2020.

\bibitem{zhao_lidar-camera_2019}
C.~Zhao, C.~Wang, B.~Zheng, J.~Hu, X.~Hou, Q.~Pan, and Z.~Xu, ``Lidar-camera
  {Based} {3D} {Obstacle} {Detection} for {UGVs},'' in \emph{{IEEE}
  {International} {Conference} on {Control} and {Automation} ({ICCA})}, 2019.

\bibitem{Survey_3D_object_detection}
E.~Arnold, O.~Y. Al-Jarrah, M.~Dianati, S.~Fallah, D.~Oxtoby, and
  A.~Mouzakitis, ``A survey on 3d object detection methods for autonomous
  driving applications,'' \emph{IEEE Transactions on Intelligent Transportation
  Systems}, vol.~20, 2019.

\bibitem{Lambert2020MSegAC}
J.~Lambert, Z.~Liu, O.~Sener, J.~Hays, and V.~Koltun, ``Mseg: A composite
  dataset for multi-domain semantic segmentation,'' \emph{2020 IEEE/CVF
  Conference on Computer Vision and Pattern Recognition (CVPR)}, 2020.

\bibitem{Lin2014MicrosoftCC}
T.-Y. Lin, M.~Maire, S.~J. Belongie, J.~Hays, P.~Perona, D.~Ramanan,
  P.~Doll{\'a}r, and C.~L. Zitnick, ``Microsoft coco: Common objects in
  context,'' in \emph{ECCV}, 2014.

\bibitem{Zhou2018SemanticUO}
B.~Zhou, H.~Zhao, X.~Puig, S.~Fidler, A.~Barriuso, and A.~Torralba, ``Semantic
  understanding of scenes through the ade20k dataset,'' \emph{International
  Journal of Computer Vision}, vol. 127, 2018.

\bibitem{Mapillary}
G.~Neuhold, T.~Ollmann, S.~R. Bulò, and P.~Kontschieder, ``The mapillary
  vistas dataset for semantic understanding of street scenes,'' in \emph{2017
  IEEE International Conference on Computer Vision (ICCV)}, 2017.

\bibitem{Varma2019IDDAD}
G.~Varma, A.~Subramanian, A.~M. Namboodiri, M.~Chandraker, and C.~V. Jawahar,
  ``Idd: A dataset for exploring problems of autonomous navigation in
  unconstrained environments,'' \emph{2019 IEEE Winter Conference on
  Applications of Computer Vision (WACV)}, 2019.

\bibitem{Yu2020BDD100KAD}
F.~Yu, H.~Chen, X.~Wang, W.~Xian, Y.~Chen, F.~Liu, V.~Madhavan, and T.~Darrell,
  ``Bdd100k: A diverse driving dataset for heterogeneous multitask learning,''
  \emph{2020 IEEE/CVF Conference on Computer Vision and Pattern Recognition
  (CVPR)}, 2020.

\bibitem{Cityscapes}
M.~Cordts, M.~Omran, S.~Ramos, T.~Rehfeld, M.~Enzweiler, R.~Benenson,
  U.~Franke, S.~Roth, and B.~Schiele, ``The cityscapes dataset for semantic
  urban scene understanding,'' \emph{2016 IEEE Conference on Computer Vision
  and Pattern Recognition (CVPR)}, 2016.

\bibitem{SUN_RGBD}
S.~Song, S.~P. Lichtenberg, and J.~Xiao, ``Sun rgb-d: A rgb-d scene
  understanding benchmark suite,'' in \emph{2015 IEEE Conference on Computer
  Vision and Pattern Recognition (CVPR)}, 2015.

\bibitem{Sun2019HighResolutionRF}
K.~Sun, Y.~Zhao, B.~Jiang, T.~Cheng, B.~Xiao, D.~Liu, Y.~Mu, X.~Wang, W.~Liu,
  and J.~Wang, ``High-resolution representations for labeling pixels and
  regions,'' \emph{arXiv: 1904.04514}, 2019.

\bibitem{Wilddash}
O.~Zendel, K.~Honauer, M.~Murschitz, D.~Steininger, and G.~F. Dominguez,
  ``Wilddash - creating hazard-aware benchmarks,'' in \emph{Proceedings of the
  European Conference on Computer Vision (ECCV)}, 2018.

\bibitem{fischlerRandomSampleConsensus1981}
M.~A. Fischler and R.~C. Bolles, ``Random sample consensus: A paradigm for
  model fitting with applications to image analysis and automated
  cartography,'' \emph{Communications of the ACM}, vol.~24, 1981.

\bibitem{edelsbrunnerShapeSetPoints1983}
H.~Edelsbrunner, D.~Kirkpatrick, and R.~Seidel, ``On the shape of a set of
  points in the plane,'' \emph{IEEE Transactions on Information Theory},
  vol.~29, 1983.

\bibitem{esterDensityBasedAlgorithmDiscovering}
M.~Ester, H.-P. Kriegel, J.~Sander, and X.~Xu, ``A density-based algorithm for
  discovering clusters in large spatial databases with noise,'' in
  \emph{Proceedings of the Second International Conference on Knowledge
  Discovery and Data Mining}.\hskip 1em plus 0.5em minus 0.4em\relax AAAI
  Press, 1996.

\bibitem{yinCenterbased3DObject2021}
T.~Yin, X.~Zhou, and P.~Krahenbuhl, ``Center-based 3d object detection and
  tracking,'' in \emph{Proceedings of the IEEE/CVF Conference on Computer
  Vision and Pattern Recognition (CVPR)}, 2021.

\bibitem{zhouVoxelNetEndtoEndLearning2018}
Y.~Zhou and O.~Tuzel, ``{{VoxelNet}}: {{End-to-End Learning}} for {{Point Cloud
  Based 3D Object Detection}},'' in \emph{2018 {{IEEE}}/{{CVF Conference}} on
  {{Computer Vision}} and {{Pattern Recognition}}}.\hskip 1em plus 0.5em minus
  0.4em\relax {IEEE}, 2018.

\bibitem{radford_learning_2021}
A.~Radford, J.~W. Kim, C.~Hallacy, A.~Ramesh, G.~Goh, S.~Agarwal, G.~Sastry,
  A.~Askell, P.~Mishkin, J.~Clark, G.~Krueger, and I.~Sutskever, ``Learning
  {Transferable} {Visual} {Models} {From} {Natural} {Language} {Supervision}.''
  in \emph{International {Conference} on {Machine} {Learning} ({ICML})}, 2021.

\bibitem{he_deep_2016}
K.~He, X.~Zhang, S.~Ren, and J.~Sun, ``Deep {Residual} {Learning} for {Image}
  {Recognition}.'' in \emph{{IEEE} {Conference} on {Computer} {Vision} and
  {Pattern} {Recognition} ({CVPR})}, 2016.

\bibitem{dosovitskiy_image_2021}
A.~Dosovitskiy, L.~Beyer, A.~Kolesnikov, D.~Weissenborn, X.~Zhai,
  T.~Unterthiner, M.~Dehghani, M.~Minderer, G.~Heigold, S.~Gelly, J.~Uszkoreit,
  and N.~Houlsby, ``An {Image} is {Worth} 16x16 {Words}: {Transformers} for
  {Image} {Recognition} at {Scale}.'' in \emph{International {Conference} on
  {Learning} {Representations} ({ICLR})}, 2021.

\bibitem{russakovsky_imagenet_2015}
O.~Russakovsky, J.~Deng, H.~Su, J.~Krause, S.~Satheesh, S.~Ma, Z.~Huang,
  A.~Karpathy, A.~Khosla, M.~Bernstein, A.~C. Berg, and L.~Fei-Fei,
  ``{ImageNet} {Large} {Scale} {Visual} {Recognition} {Challenge},''
  \emph{International Journal of Computer Vision}, vol. 115, 2015.

\bibitem{carion_end--end_2020}
N.~Carion, F.~Massa, G.~Synnaeve, N.~Usunier, A.~Kirillov, and S.~Zagoruyko,
  ``End-to-{End} {Object} {Detection} with {Transformers}.'' in \emph{European
  {Conference} on {Computer} {Vision} ({ECCV})}, 2020.

\bibitem{lin_microsoft_2014}
T.-Y. Lin, M.~Maire, S.~J. Belongie, J.~Hays, P.~Perona, D.~Ramanan,
  P.~Dollár, and C.~L. Zitnick, ``Microsoft {COCO}: {Common} {Objects} in
  {Context}.'' in \emph{European {Conference} {Computer} {Vision} ({ECCV})},
  2014.

\bibitem{guo_calibration_2017}
C.~Guo, G.~Pleiss, Y.~Sun, and K.~Q. Weinberger, ``On {Calibration} of {Modern}
  {Neural} {Networks}.'' in \emph{International {Conference} on {Machine}
  {Learning} ({ICML})}, 2017.

\bibitem{nguyen_deep_2015}
A.~M. Nguyen, J.~Yosinski, and J.~Clune, ``Deep neural networks are easily
  fooled: {High} confidence predictions for unrecognizable images.'' in
  \emph{{IEEE} {Conference} on {Computer} {Vision} and {Pattern} {Recognition}
  ({CVPR})}, 2015.

\bibitem{hendrycks_baseline_2017}
D.~Hendrycks and K.~Gimpel, ``A {Baseline} for {Detecting} {Misclassified} and
  {Out}-of-{Distribution} {Examples} in {Neural} {Networks}.'' in
  \emph{International {Conference} on {Learning} {Representations} ({ICL})},
  2017.

\bibitem{minderer_revisiting_2021}
M.~Minderer, J.~Djolonga, R.~Romijnders, F.~Hubis, X.~Zhai, N.~Houlsby,
  D.~Tran, and M.~Lucic, ``Revisiting the {Calibration} of {Modern} {Neural}
  {Networks},'' \emph{arXiv:2106.07998}, 2021.

\bibitem{liInsClusteringInstantlyClustering2020}
Y.~Li, C.~L. Bihan, T.~Pourtau, and T.~Ristorcelli, ``Insclustering: Instantly
  clustering lidar range measures for autonomous vehicle,'' in \emph{2020 IEEE
  23rd International Conference on Intelligent Transportation Systems (ITSC)},
  2020.

\bibitem{li_coda_2022}
K.~Li, K.~Chen, H.~Wang, L.~Hong, C.~Ye, J.~Han, Y.~Chen, W.~Zhang, C.~Xu,
  D.-Y. Yeung, X.~Liang, Z.~Li, and H.~Xu, ``{CODA}: {A} {Real}-{World} {Road}
  {Corner} {Case} {Dataset} for {Object} {Detection} in {Autonomous}
  {Driving},'' \emph{arXiv:2203.07724}, 2022.

\bibitem{geiger_are_2012}
A.~Geiger, P.~Lenz, and R.~Urtasun, ``Are we ready for autonomous driving?
  {The} {KITTI} vision benchmark suite.'' in \emph{{IEEE} {Conference} on
  {Computer} {Vision} and {Pattern} {Recognition} ({CVPR})}, 2012.

\bibitem{caesar_nuscenes_2020}
H.~Caesar, V.~Bankiti, A.~H. Lang, S.~Vora, V.~E. Liong, Q.~Xu, A.~Krishnan,
  Y.~Pan, G.~Baldan, and O.~Beijbom, ``{nuScenes}: {A} {Multimodal} {Dataset}
  for {Autonomous} {Driving}.'' in \emph{{IEEE} {Conference} on {Computer}
  {Vision} and {Pattern} {Recognition} ({CVPR})}, 2020.

\bibitem{mao_one_2021}
J.~Mao, M.~Niu, C.~Jiang, H.~Liang, J.~Chen, X.~Liang, Y.~Li, C.~Ye, W.~Zhang,
  Z.~Li, J.~Yu, H.~Xu, and C.~Xu, ``One {Million} {Scenes} for {Autonomous}
  {Driving}: {ONCE} {Dataset},'' \emph{arXiv:2106.11037}, 2021.

\end{thebibliography}
}

\end{document}